\documentclass[runningheads]{llncs}
\usepackage{graphicx}
\usepackage{hyperref}
\hypersetup{
    colorlinks=true,
    linkcolor=blue,
    filecolor=magenta,      
    urlcolor=cyan,
}

\usepackage{amsmath,amssymb} % define this before the line numbering.
\usepackage{color}

\usepackage{nameref}

\usepackage[usenames,dvipsnames,svgnames,table]{xcolor}
\usepackage{xspace}
\usepackage[normalem]{ulem}

\makeatletter
\DeclareRobustCommand\onedot{\futurelet\@let@token\@onedot}
\def\@onedot{\ifx\@let@token.\else.\null\fi\xspace}
\def\eg{\emph{e.g}\onedot} 

\def\ie{\emph{i.e}\onedot}

\def\etal{\emph{et al}\onedot}
\makeatother

\newcommand{\expect}{\mathbb{E}}

\usepackage[capitalize]{cleveref}
\renewcommand{\vec}[1]{\boldsymbol{\mathrm{#1}}}

\usepackage{float}

\begin{document}
\title{A Differentiable Distance Approximation for Fairer Image Classification}
\author{Nicholas Rosa\inst{1} \and
Tom Drummond\inst{1,2} \and
Mehrtash Harandi\inst{1}}
\authorrunning{N. Rosa et al.}
\institute{Monash University,  Australia \and
The University of Melbourne, Australia
}
\maketitle              % typeset the header of the contribution
\begin{abstract}
Na\"{\i}vely trained AI models can be heavily biased. This can be particularly problematic when the biases involve legally or morally protected attributes such as ethnic background, age or gender. Existing solutions to this problem come at the cost of extra computation, unstable adversarial optimisation or have losses on the feature space structure that are disconnected from fairness measures and only loosely generalise to fairness. In this work we propose a differentiable approximation of the variance of demographics, a metric that can be used to measure the bias, or unfairness, in an AI model. Our approximation can be optimised alongside the regular training objective which eliminates the need for any extra models during training and directly improves the fairness of the regularised models. We demonstrate that our approach improves the fairness of AI models in varied task and dataset scenarios, whilst still maintaining a high level of classification accuracy. Code is available at \url{https://bitbucket.org/nelliottrosa/base_fairness}.
\end{abstract}

\section{Introduction} \label{intro}

In recent times, the use of Artificial Intelligence (AI) has permeated many processes that are used to make important decisions, such as filtering applicants for jobs, deciding if an applicant should receive credit and recognizing people in images~\cite{schroff_facenet_2015,joseph_fairness_2016-1}. Given this, it is essential to ensure that  AI-driven models are not exhibiting behaviour which is morally or legally undesirable.
In AI, data is a collection of attributes, which can either be explicit (\eg labels) or implicit (\eg information from an image). Some of these attributes are referred to as \emph{protected} attributes as they should not be used to discriminate (\eg gender, race or age). However, it has been shown numerous times that AI models which are na\"{\i}vely trained are biased against one or more of these protected attributes, as they exhibit lower accuracy for some demographics \cite{buolamwini_gender_2018,grother_face_2019,klare_face_2012}. This behaviour is discriminatory against these demographics and is \emph{morally or legally undesirable}, or simply unfair.
There are two common sources of unfair behaviour that can present itself in AI systems. The first source is biases that are present in the data used for training AI models. Biases in the data with respect to protected attributes can cause an AI model trained upon that data to discriminate against the protected attribute \cite{beutel_data_2017}. For example, if a dataset used to train a facial recognition model for unlocking doors, only contains images of men (\ie bias with respect to gender) then the learned model will not accurately recognise and admit women (\ie unfair behaviour). The second source of bias is due to some values or demographics of a protected attribute being inherently harder for AI to recognize than others. For example, it has been shown that even when training with a balanced dataset, faces with a darker skin tone are harder to recognize for facial recognition algorithms \cite{wang_mitigating_2020}.

Various solutions to the fairness problem have been proposed. We focus on algorithmic in-processing methods for reducing the bias\cite{vedaldi_jointly_2020,zhang_mitigating_2018,creager_flexibly_2019,wang_balanced_2019,dhar_pass_2021,gong_mitigating_2021,cho_fair_2020,hwang_exploiting_2020}. In-processing aims to address the bias of a model by applying an extra objective during training which makes the model bias-aware and consequentially learns a fairer model. In-processing has proven to be quite effective at reducing the unfair behaviour of AI. However, in-processing methods often include extra models which can increase training cost and complexity \cite{jung_fair_2021}; use adversarial training \cite{vedaldi_jointly_2020,zhang_mitigating_2018,dhar_pass_2021,creager_flexibly_2019} which has proven to be notoriously unstable \cite{roth_stabilizing_2017} or make assumptions about the representation space of the model which may not hold in all cases\cite{gong_mitigating_2021,park_fair_2022,xu_consistent_2021}.
Creating fair AI models is particularly difficult in the computer vision domain as any problems with extra computational cost and complexity are exacerbated by the large models utilized. Additionally the high dimensionality of images means they can contain many implicit attributes, which are often highly correlated to each other and to protected attributes. Disentangling the implicit factors is extra challenging in these cases. 

In this paper, we introduce \textbf{B}ias \textbf{A}ccuracy \textbf{S}tandard deviation \textbf{E}stimation or BASE, a novel fairness algorithm, which optimizes a differentiable approximation of the fairness metric \emph{standard deviation of accuracy across demographics} ($\sigma_{acc}$) to learn an AI model which is fair with respect to \emph{equalized odds} (EO). Models that exhibit a low standard deviation of accuracy across demographics or variance of demographics have the property of equal performance on a target task regardless of the demographic of the protected attribute. For example, a facial recognition model which has low variance of demographics for ethnicity, is equally likely to correctly recognize the identity of a person from an image regardless of their ethnicity. Reducing the variance of demographics of a model makes it fairer w.r.t. EO. However, for an AI model that is trained with gradient based optimization the variance of demographics is difficult to use. This is due to the accuracy of a single sample - an integral part of the variance of demographics (\cref{ssec:metrics}) - having an undefined gradient at 0 and being 0 everywhere else, which leads to zero influence on the model parameters. BASE overcomes this difficulty by instead using a sigmoid based approximation of accuracy which we call \emph{soft-accuracy} inside the variance of demographics metric. This approach has multiple advantages. Firstly computational efficiency, for example, training a classifier on images with BASE incurs only the extra computation of calculating the variance of demographics. Compare this to training a classifier with knowledge distillation \cite{jung_fair_2021} or adversarial debiasing \cite{zhang_mitigating_2018}, where additional models are used which incur extra memory usage for the model parameters and gradients, alongside with extra computation for the forward pass of the additional model. Secondly, BASE makes no assumptions about the representation space. The model will automatically learn the representation space structure required to reduce the variance of demographics. Furthermore, due to its simplicity BASE can be combined with other solutions.

To summarize the main contributions of our work are:
\begin{itemize}
    \item Provide a novel method for improving the fairness on AI models trained with gradient based optimization, that increases algorithmic simplicity and does not rely on training additional models (\cref{ssec:diff}).
    \item Show that our method is competitive with and in some cases outperforms current state-of-the-art fair image classifiers when using either a biased dataset or an unbiased dataset (\cref{sssec:unbalanced}, \cref{sssec:balanced}).
    \item Show that our method increasingly outperforms the fairness of a naive classifier when exposed to increasingly biased training sets in which target and protected attributes are strongly correlated. Our method also achieves higher over-all accuracy on heavily biased datasets (\cref{sssec:biased_data}).
\end{itemize}

% \MH{OK, the intro needs a lot of work. You do not have much to talk about your algorithm yet. My preference to write the intro is to start by stating what you have done in the paper. That is, ``In this paper, we propose a differentiable ....'' Then you can say, why it is important to do this (like the first few sentences at the beginning of the intro right now. instead of $\sigma_{\mathrm{acc}}$, you can say variance of demographics (I suppose). you can explain what would be the property of a fair model wrt $\sigma_{\mathrm{acc}}$ and why it is difficult to use the metric $\sigma_{\mathrm{acc}}$ directly (be explicit). When it comes to the pros (computational load and etc), be more explicit. Give exmaples, ``For example, training X with BIAS incurs only Y . Also, your method can be combined with other solutions verbatim (again I suppose). I think this is a very big plus. if that is the case, advertise it }
\section{Related Work and Preliminaries}

Fair AI has received increasing attention in the past few years and a varied range of solutions has been proposed. Algorithmic methods for reducing the bias can be broken down into three main categories based upon when they apply their fairness constraint. \emph{Pre}-processing methods aim to change the distribution of the data used for training such that a fairer model is produced. These methods include re-sampling, which changes the sampling rate of data during training to ensure each protected class is equally represented \cite{amini_uncovering_2019,roh_fairbatch_2021,shekhar_adaptive_2021} and augmentation methods which add synthetic data to the dataset \cite{ramaswamy_fair_2021,xu_fairgan_2018,van_breugel_decaf_2021,zietlow_leveling_2022} to balance the protected classes. The second class of methods, \emph{post}-processing methods, aim to adjust the prediction after the fact to compensate for the bias \cite{wang_towards_2020}. Pre-processing and post-processing have some major drawbacks. Pre-processing only addresses the bias in the dataset and the inherent difficulties of some demographics can still cause a biased model \cite{wang_balanced_2019,wang_mitigating_2020}.
On the other hand post-processing methods require that protected attribute labels to be known at inference time or assume that the target and protected attribute are independent \cite{wang_towards_2020}. Our method is related to the final category of \emph{in}-processing, which is discussed further below. In-processing methods typically run under a constrained optimization scheme where a loss penalty or a special construction of the AI model is used to reduce the bias during optimization.

\subsection{In-processing for Fair Classification}

Like many machine learning tasks, the fairness problem is difficult to optimize directly and adversarial training became a common method to create fair representations and predictors \cite{vedaldi_jointly_2020,zhang_mitigating_2018,creager_flexibly_2019,wang_balanced_2019,dhar_pass_2021}. These methods use an adversarial model, or adversary, whose purpose is to learn the relationship between the predictor and the protected attribute. The output of the adversary is then used to enforce a fairness constraint upon the predictor. This is achieved either by gradient reversal of the adversary or by maximising the entropy of the adversaries predictions. If a strong adversary is unable to determine a relationship between the predictor and the protected attribute then fairness of the predictor can be guaranteed \cite{zhang_mitigating_2018}

Other constrained optimization methods have been proposed and their approaches vary greatly. Gong \etal \cite{gong_mitigating_2021} minimize the variance of sample density across different demographics within the representation space. Cho \etal \cite{cho_fair_2020} use a kernel density estimate to approximate the conditional distributions used for measuring fairness in a differentiable manner. Hwang \etal \cite{hwang_exploiting_2020} reduce the Wasserstein distance between protected groups within the representation space. Finally, in a work most similar to our own Shen \etal \cite{shen_optimising_2022} use cross-entropy loss as a proxy for probability during training to optimise for fairness. Our method differs in two main aspects; our objective directly considers the two elements of the models output vector responsible for determining accuracy and we evaluate our work in the computer vision domain.

\subsection{Problem definition}
The ultimate goal of fair machine learning is to create predictors which contain no bias. There is, however, many different forms of bias that can present themselves and as a consequence there are multiple different definitions of fairness. The three most common definitions are Demographic parity \cite{zhang_mitigating_2018}, Equalized Odds \cite{hardt_equality_2016} and Equalized Opportunity \cite{hardt_equality_2016}. In the following section $A$, $\hat{Y}$ and $Y$ are random variables which represent the protected attribute, the output of a predictor and the true value of the target attribute respectively.

\subsubsection{Demographic Parity}
Demographic parity is the simplest form of fairness since it only considers the output of the predictor and the protected attribute. A predictor satisfies demographic parity when its output is independent of the protected attribute. That is $\forall a\in \mathcal{A}; \mathrm{Pr}(\hat{Y}=\hat{y}|A=a) = \mathrm{Pr}(\hat{Y}=\hat{y})$. However, this definition does not always allow for perfect classification \cite{hardt_equality_2016}. If there is any correlation between the protected attribute and the target task then maintaining independence forces a reduction in performance. For example, if we learned a predictor for university admittance with age as a protected attribute, then achieving demographic parity would require our predictor to admit young children with the same probability as those who had just finished high school, regardless of each individuals suitability.

\subsubsection{Equalized Odds}
Equalized Odds is another definition of fairness that is more commonly applied for computer vision tasks. A predictor satisfies equalized odds when its output is conditionally independent of the protected attribute for all classes of the target class. That is  $\forall y\in \mathcal{Y}, \forall a,a'\in \mathcal{A}, \mathrm{Pr}(\hat{Y}=y|A=a,Y=y) = \mathrm{Pr}(\hat{Y}=y|A=a',Y=y)$. 
This definition allows us to maintain performance as it is satisfied when a predictor achieves the same level of accuracy for each demographic of the protected attribute.

\subsubsection{Equalized Opportunity}
Equalized Opportunity is a special case of equalized odds for which there is a class of the target task $y_+\in \mathcal{Y}$ that confers advantage, \eg, to receive a loan or be hired for a job. It is a relaxation of equalized odds that is satisfied when the output of the predictor is conditionally independent of the protected attribute for only the advantageous class. That is $\forall a,a'\in \mathcal{A}, \mathrm{Pr}(\hat{Y}=y_+|A=a,Y=y_+) = \mathrm{Pr}(\hat{Y}=y_+|A=a',Y=y_+)$
\\\\
Equalized odds and equalized opportunity are more practical definitions of fairness when applied to a computer vision problems because they still allow full predictive capability \cite{hardt_equality_2016}. Further, since equalized opportunity is a relaxation of equalized odds, if equalized odds is achieved then equalized opportunity is also achieved. Therefore, in this work we aim to create predictors that satisfy equalized odds. 

\subsection{Distance measures for Equalized Odds}\label{ssec:metrics}
Though the goal is to achieve true equalized odds, current methods are unable to achieve it \cite{jung_fair_2021,zhang_mitigating_2018,cho_fair_2020}. Therefore, we need to use metrics to quantify how far a predictor is from true equalized odds. In this work we use three different metrics to measure the level of fairness of a predictor. The first two metrics use the difference in predictor output between different demographics of a protected attribute. This difference is called the \emph{difference of equalized odds} (DEO). 

\begin{equation}\label{deo}
    \mathrm{DEO}(a,a',y) \triangleq \big|\mathrm{Pr}(\hat{Y}=y|A=a,Y=y) - \mathrm{Pr}(\hat{Y}=y|A=a',Y=y)\big|\;.
\end{equation}

DEO can be directly used when the protected attribute is binary and can easily be extended for more demographics by aggregating DEO across the different target and protected attribute values.
The methods used to aggregate DEO differ between various works in the literature. We use the aggregation methods from Jung \emph{et al.} \cite{jung_fair_2021} who propose two different methods of aggregation, $\mathrm{DEO_{max}}$ and $\mathrm{DEO_{avg}}$ which are shown in \cref{deom,deoa}, respectively. $\mathrm{DEO_{max}}$ can be used to understand the peak bias of an AI model and $\mathrm{DEO_{avg}}$ can be used to understand the bias of a model in the majority of cases.

\begin{equation}\label{deom}
    \mathrm{DEO_{max}} \triangleq \max_{y}(\max_{a,a'}(\mathrm{DEO}(a,a',y)))\;.
\end{equation}

\begin{equation}\label{deoa}
    \mathrm{DEO_{avg}} \triangleq \frac{1}{|\mathcal{Y}|}\sum_{y}(\max_{a,a'}(\mathrm{DEO}(a,a',y)))\;.
\end{equation}

Another fairness metric that is commonly reported, often in the Fair face recognition literature, is the standard deviation of accuracy across the demographics of the protected attribute, denoted by $\sigma_{\mathrm{Acc}}$. This metric is shown in \cref{acc_std}, where $\mu$ is the average accuracy across all the demographics. Note that $\mathrm{Pr}(\hat{Y}=y|A=a)$ is equivalent to the accuracy of the predictor $\hat{Y}$ in the domain of demographic $a$.

\begin{equation}\label{mu}
    \mu = \frac{1}{|\mathcal{A}|}\sum_{a\in\mathcal{A}}[\mathrm{Pr}(\hat{Y}=y|A=a)]
\end{equation}

\begin{equation}\label{acc_std}
    \sigma_{\mathrm{Acc}} \triangleq \sqrt{\frac{1}{|\mathcal{A}|}\sum_{a}\left[\mathrm{Pr}(\hat{Y}=y|A=a) - \mu\right]^2}
\end{equation}

All these metrics represent a distance from true equalized odds. In all cases this means that lower values indicate a fairer classifier.
\section{Method}

\subsection{A differentiable approximation for distance from Equalized Odds}\label{ssec:diff}
The strategy used to train an AI model for classification uses a distance measure between the models output distribution and the true data distribution, referred to as the \textit{loss} or \textit{objective} function. Then a gradient optimization method is used to update the parameters of the model to reduce the distance measure. This is a simple but incredibly effective strategy. We aim to use the same strategy to increase the fairness of an AI model. We use $\sigma_{\mathrm{Acc.}}$ as an objective function to reduce the distance from true EO. 

In what follows, we use boldface fonts to denote vectors, \eg, $\vec{\hat{y}} \in \mathcal{\hat{Y}}$ denotes the output vector of the model. We use $\hat{y}_t$ to show the element  corresponding to the ground truth label $y$ in $\vec{\hat{y}}$. Furthermore, $\hat{y}_m = \mathrm{max}(\vec{\hat{y}}\setminus\{\hat{y}_{t}\})$ represents the largest non ground truth element of $\vec{\hat{y}}$ and $\mathcal{\hat{Y}}_a$ represents the domain of demographic $a$ for the protected attribute.
Accuracy of a single sample $\vec{\hat{y}}$ is defined in \cref{acc}. 

\begin{equation}\label{acc}
    \mathrm{Acc}(\hat{y}_t, \hat{y}_m) \triangleq
    \begin{cases} 
        1 & \hat{y}_t > \hat{y}_m \\
        0 & \mathrm{otherwise.}
    \end{cases}
\end{equation}

In essence, if the element $\hat{y}_t$ is greater than all other elements, the model has correctly predicted the outcome for this sample and therefore, has an accuracy of one.

Since $\expect_{\vec{\hat{y}}\sim\mathcal{\hat{Y}}_a}[\mathrm{Acc}(\hat{y}_t, \hat{y}_m)]=\mathrm{Pr}(\hat{Y}=y|A=a)$, we substitute the expectation into \cref{mu,acc_std}, which gives us \cref{mu_expect,acc_std_exp}. 

\begin{equation}\label{mu_expect}
    \mu = \frac{1}{|\mathcal{A}|}\sum_{a\in\mathcal{A}}\expect_{\vec{\hat{y}}\sim\mathcal{\hat{Y}}_a}[\mathrm{Acc}(\hat{y}_t, \hat{y}_m)]
\end{equation}

\begin{equation}\label{acc_std_exp}
    \sigma_{\mathrm{Acc}} = \sqrt{\frac{1}{|\mathcal{A}|}\sum_{a\in\mathcal{A}}\left[\expect_{\vec{\hat{y}}\sim\mathcal{\hat{Y}}_a}[\mathrm{Acc}(\hat{y}_t, \hat{y}_m)] - \mu\right]^2}
\end{equation}

This is the objective we would like to optimize. However to be used for gradient based optimization that AI models are trained with an objective needs to be differentiable, which $\sigma_{\mathrm{Acc}}$ is not due to the undefined gradient at $\hat{y}_t=\hat{y}_m$ of $\mathrm{Acc}(\hat{y}_t, \hat{y}_m)$. Instead we approximate the accuracy using a sigmoid based soft accuracy function, shown in \cref{soft_acc}, which is a differentiable approximation of accuracy. The soft accuracy is characterised by $\kappa$, which is a hyper-parameter that describes the sharpness of the function. A higher value of $\kappa$ leads to a closer approximation of accuracy with $\lim_{\kappa\to\infty} \mathrm{Acc_{soft}}(\hat{y}_t, \hat{y}_m) = \mathrm{Acc}(\hat{y}_t,\hat{y}_m)$, however this is paired with an increased sparsity of the gradient.

\begin{equation}\label{soft_acc}
    \mathrm{Acc_{soft}}(\hat{y}_t, \hat{y}_m) \triangleq \frac{1}{1+e^{-\kappa(\hat{y}_t-\hat{y}_m)}}
\end{equation}

We then substitute soft accuracy into $\sigma_{Acc}$ for accuracy. This gives us the objective shown in \cref{sacc_std}.

\begin{equation}\label{mu_soft}
    \mu_{\mathrm{soft}} = \frac{1}{|\mathcal{A}|}\sum_{a\in\mathcal{A}}\expect_{\vec{\hat{y}}\sim\mathcal{\hat{Y}}_a}[\mathrm{Acc_{soft}}(\hat{y}_t, \hat{y}_m)]
\end{equation}

\begin{equation}\label{sacc_std}
    \sigma_{\mathrm{Acc_{soft}}} \triangleq \sqrt{\frac{1}{|\mathcal{A}|}\sum_{a\in\mathcal{A}}\Big[\expect_{\vec{\hat{y}}\sim\mathcal{\hat{Y}}_a}\big[ \mathrm{Acc_{soft}}(\hat{y}_t, \hat{y}_m)\big] - \mu_{\mathrm{soft}}\Big]^2}
\end{equation}

This is the differentiable objective that we can optimize to obtain a fair predictor.

\subsection{Training objective}
By itself the soft accuracy fairness objective does not learn to classify. In fact the easiest solution for a model to achieve equalized odds is to randomly classify each sample. Since it is important that the model still achieves high utility we combine the soft accuracy fairness objective with a cross entropy classification objective. This gives us the full objective which is shown in \cref{obj}.

\begin{equation}\label{obj}
    \mathcal{L} = \mathcal{L}_{\mathrm{ce}} + \gamma \sigma_{\mathrm{Acc_{soft}}}
\end{equation}

The two losses, $\mathcal{L}_{ce}$ and $\sigma_{\mathrm{Acc_{soft}}}$, aim to achieve different objectives, which are classification performance and fairness respectively. Applying too much weight to one objective can harm the other. We use $\gamma$ as a hyper-parameter to balance the utility of the model with the fairness. A higher value for $\gamma$ will result in a fairer classifier, however at this can often come at the cost of classification performance. We experimentally determined the optimal value of $\gamma$ for each dataset by performing a grid search. However, we observe an extensive search is not required and finding the correct order of magnitude results in good performance.

\subsection{Balancing the training dataset}
When calculating $\sigma_{\mathrm{Acc_{soft}}}$ on a mini-batch the number of samples used to estimate the soft accuracy for each protected demographic is highly important. If the number of samples for a particular demographic is too low then the variance of the soft accuracy estimation will increase. Differences in variance between the different demographics lead to instability of training gradients which has a negative impact on performance. To counter this effect we simply oversample the training dataset set such that each protected, target attribute pair is evenly sampled. This is achieved by randomly duplicating samples from the undersampled pairs until all protected, target attribute pairs contain the same number of samples. There exist more sophisticated methods \cite{xu_fairgan_2018,ramaswamy_fair_2021} which could be used to augment the training dataset and their use may lead to gains in performance. However, we leave this investigation to future work.

\section{Experiments}
In the following section we thoroughly investigate and validate the capability of our soft accuracy fairness objective.

\subsection{Baselines}
We compare our algorithm with four different baselines. The first is a na\"{\i}ve classifier that is not aware of fairness in any regard. This baseline represents the worst case scenario for fairness. Since one source of bias is an unbalanced dataset we also include a na\"{\i}ve classifier which is trained by oversampling the dataset such that it is balanced. We refer to this baseline as \textit{Na\"{\i}ve Balanced}. The third baseline is Adversarial Debiasing (AD) \cite{zhang_mitigating_2018} which is used as a common benchmark method and the final baseline is the state-of-the-art in-processing method, MFD \cite{jung_fair_2021}. The original MFD paper only provided results for the age task with the UTKFace dataset. Additionally, the original MFD paper only implemented a simple data augmentation scheme. We employed further data augmentations which allowed our na\"{\i}ve classifier to achieve a much higher accuracy ($74.7\%$ vs $83.1\%$). In the spirit of fair comparison, we apply their method code with our datasets and augmentation scheme, this allows MFD to achieve comparable accuracy. Where applicable, results from the original paper are reported as MFD$^\diamond$. Similarly, AD was originally implemented on non computer vision tasks, we re-implement AD for evaluation on CV tasks. For both re-implementations we perform a sweep of the bias loss hyper-parameter, discard hyper-parameter choices that lead to a large reduction in accuracy and report the best results.

\subsection{Datasets}
\begin{figure}[!h]
\centering
\includegraphics[width=0.6\textwidth]{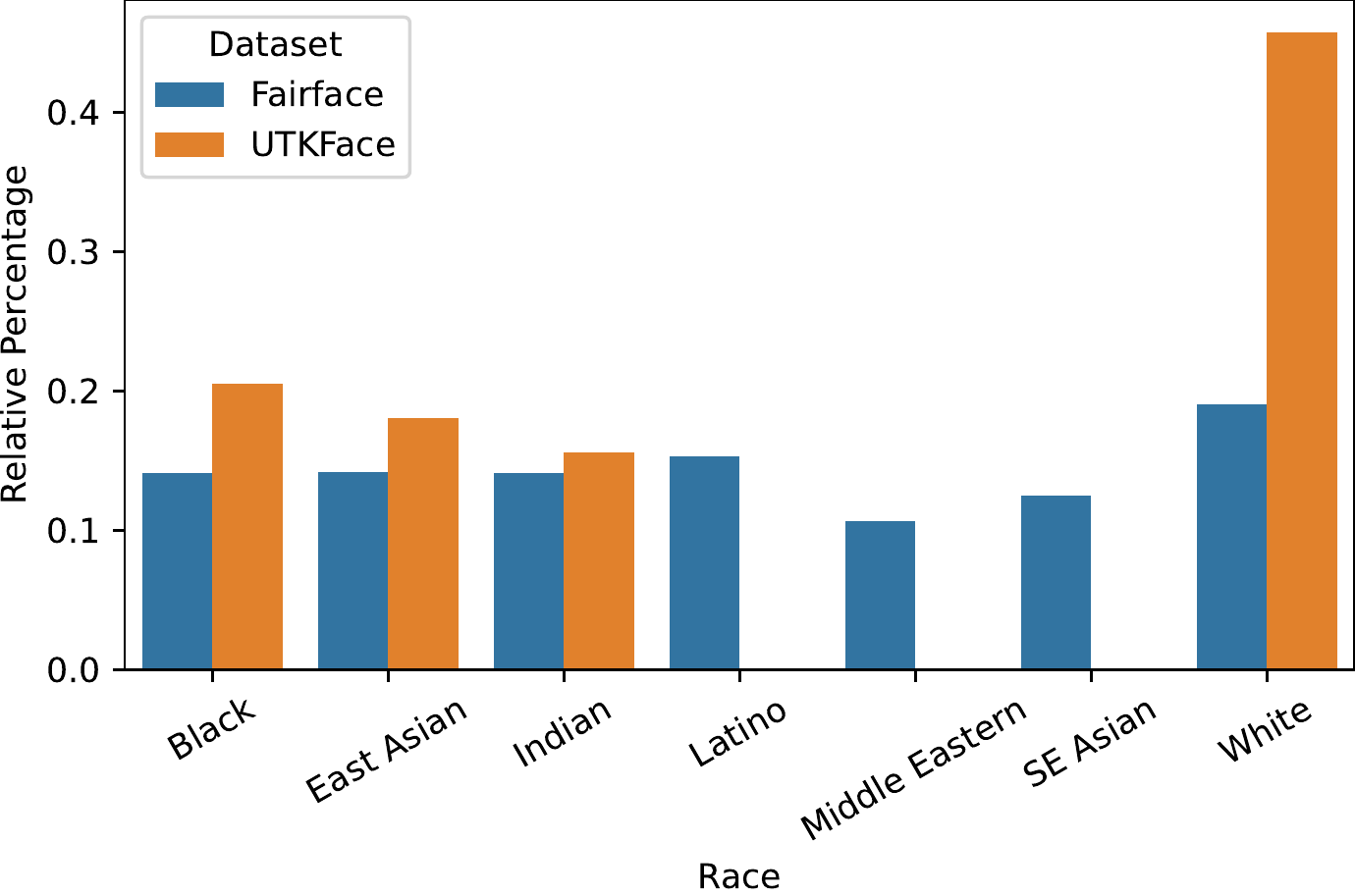}
\caption{
The relative distribution of different races in the UTKFace dataset and Fairface dataset. UTKFace labels both East Asian and South East Asian faces together so these are shown under East Asian.
}
\label{fig:race_dist}
\end{figure}

We use three datasets for our experiments. UTKFace \cite{kamiran_classifying_2009}, CelebA \cite{liu2015faceattributes} and Fairface \cite{karkkainen_fairface_2021}. UTKFace and CelebA are face image datasets commonly used to benchmark fairness. UTKFACE contains $~20\mathrm{k}$ samples with annotations of age, gender and ethnicity. CelebA contains $~200\mathrm{k}$ images which are labelled with 40 binary attributes. The images from UTKFace and CelebA cover a large variation in position, facial expression, illumination, occlusion and resolution. 
Buolamwini and Gebru \cite{buolamwini_gender_2018} note that collecting a balanced dataset should be the first step in a fairness solution. Therefore it is important that we also evaluate our method under these conditions, for which we use the Fairface dataset.
Fairface is also a face image dataset. It contains $~98\mathrm{k}$ images with annotations of age, gender and ethnicity. Fairface was created in an effort to reduce racial bias in existing datasets and had a strong focus on reducing the imbalance of races in the dataset during its creation. As shown in \cref{fig:race_dist}, compared to UTKFace, the race labels in Fairface are much more balanced. Using these UTKFace, CelebA and Fairface we evaluate two scenarios. Where a task is trained with a balanced dataset and where the task is trained with a biased dataset. UTKFace provides the age labels as integers, instead of learning a regression problem we group ages together into classes. To allow comparison we follow the division used by Jung \etal \cite{jung_fair_2021} where ages are divided into three classes, less than 20, 20-39 and greater than 40. Fairface provides age labels in classes already, however they are heavily imbalanced with far fewer samples in the extreme young and old classes. To maintain Fairface as a balanced test set we divide the ages in four new classes to balance them. These four classes are 0-19, 20-29, 30-39 and 40+.

\subsubsection{Skewed Fairface}\label{fairfaces}
\begin{figure}[!h]
\centering
\includegraphics[width=0.6\textwidth]{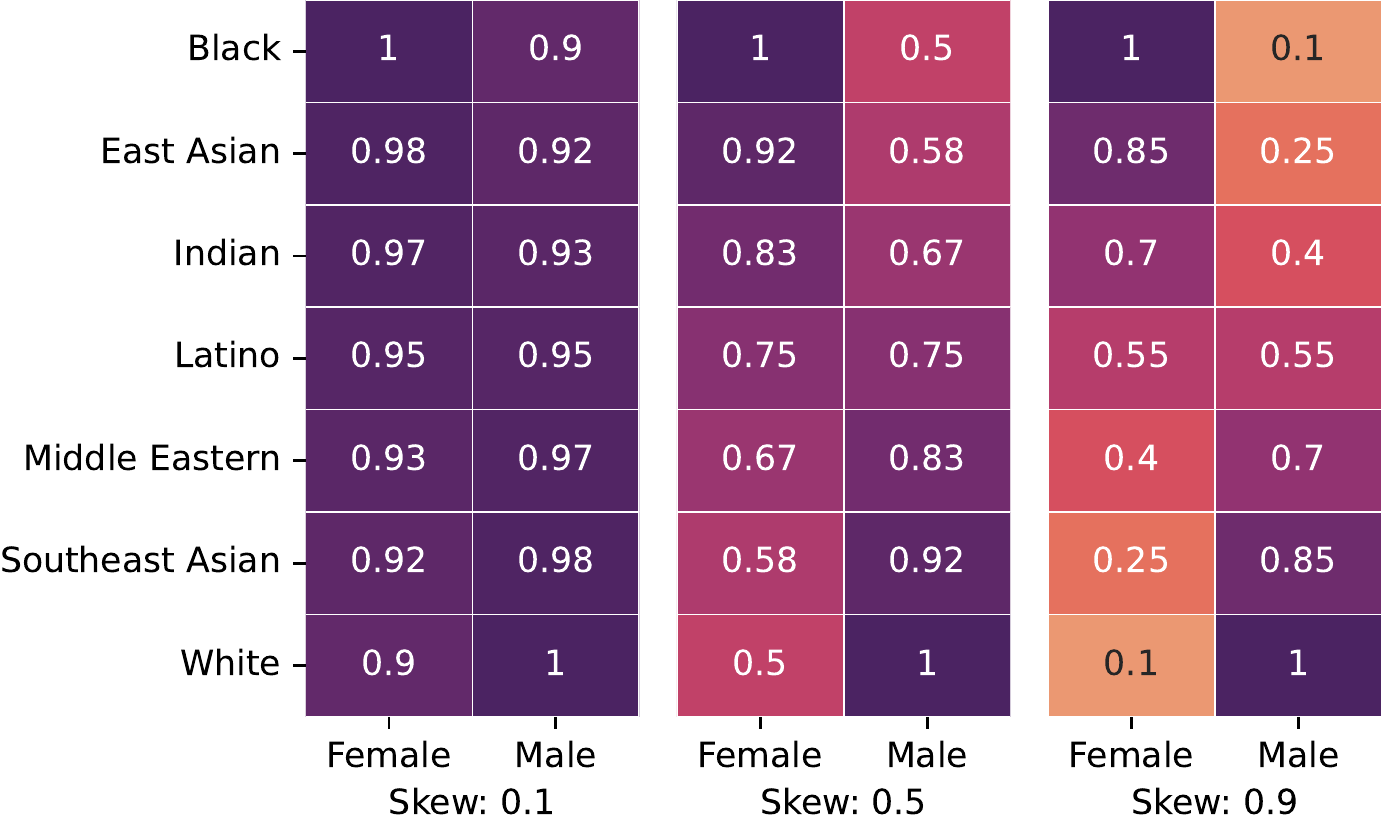}
\caption{
The relative distribution of (protected, target) pairs in the FairfaceS dataset for different skew values.
}
\label{fig:skew}
\end{figure}

Since it is imperative to understand how the performance of a fairness algorithm is related to the bias of a dataset, we present a protocol for controlling the bias within a dataset. We apply this protocol to Fairface to create a dataset which we name Fairface Skewed (FairfaceS). FairfaceS is characterised by a skew parameter ($s$) which can range from 0 to 1. The skew parameter describes the relative distribution of (target, protected) attribute pairs where a higher skew parameter leads to a dataset with a higher correlation between the target attribute and the protected attribute. The relative distribution is calculated by arranging the classes into a 2D array. Two diagonal corners are assigned the value of 1 and the other two diagonal corners are assigned the value of $1-s$. Bilinear interpolation is then used to calculate the remaining values of the matrix. An example of the relative distribution for different skews is shown in \cref{fig:skew}. Fairface is then under-sampled such that the relative distribution of each (target, protected) pair matches that in the matrix. This protocol imposes an order on the class however, in the absence of a rigorous similarity metric between separate demographics and target attribute values we simply order the classes alphabetically. 

As the skew value increases the mutual information between the protected attribute and the target attribute increases leading to an increase in the bias. Using FairfaceS allows us to evaluate how a fairness algorithm performs under varying degrees of dataset bias. Additionally, because FairfaceS uses genuine attributes that can be linked in complicated manners, rather than creating a bias with respect to augmentations such as grayscaling an image, it allows for greater understanding of how a system may behave in a real-world scenario.

\subsubsection{Balanced Test Set and Triplicate experiments.}
When evaluating the fairness of a model it is important that the test set have a uniform distribution of (protected, target) pairs. If a particular pair is undersampled or oversampled it will have a disproportionate impact on the results, e.g. if the White, Male pair is more prevalent in the test set then the accuracy on this pair will affect the average accuracy more. To ensure that our results do not include any bias toward a particular label pair we select samples for the test set such that each protected, target label pair is included in equal numbers.
We also observed that whilst the target classification performance is stable over different training and test splits, the fairness varies by a large degree. To ensure robust results we perform our experiments on three different train test splits and report the mean and standard deviation. The exception is our experiments with CelebA, for which we use the official train, validation, and test sets as this allows us to compare to previous work. The results for CelebA are reported over three different random initializations.

\subsection{Implementation Details}
For all experiments we use a Resnet18 \cite{he_deep_2016}. For experiments on UTKFace, Fairface and FairfaceS models are initialised from weights that were pretrained on Imagenet-1k \cite{deng_imagenet_2009}. Models in the CelebA experiments are randomly initialised. More details about the exact training procedure can be found in the supplementary material. MFD and AD are implemented according to their original papers. However, we follow Jung \etal \cite{jung_fair_2021} and remove the gradient projection from the original work to increase stability of training.

\subsection{Classification Tasks}
In this section we investigate the performance of our method on two tasks, age and gender classification.

\subsubsection{Unbalanced data}\label{sssec:unbalanced}
First we test the scenario in which the training data from the task is not balanced. This is the case for the majority of AI tasks unless special care has been taken during the creation of the dataset. For this experiment, we use the UTKFace and CelebA datasets. For UTKFace we use \textit{race} as the protected attribute and test both age and gender as target attributes. For CelebA we use the \textit{Male} attribute as the protected attribute and \textit{Attractive} as the target attribute. The results are shown in tables \ref{table:utkface_age}, \ref{table:utkface_gender} and \ref{table:celeba_attractive}, respectively. In both UTKFace scenarios all fairness methods improve fairness over a na\"{\i}ve classifier. The age classification task in harder than gender and is also much less fair, with the na\"{\i}ve classifier only achieving a $\sigma_{\mathrm{Acc}}$ of $8.5$ compared to $3.0$ for the gender task. In the highly unfair scenario with age as the target attribute, we observe that BASE achieves the best fairness for $\sigma_{\mathrm{Acc}}$ and $\mathrm{DEO_{avg}}$, whilst achieving the highest over-all accuracy. It is only outperformed on $\mathrm{DEO_{max}}$ by MFD$^\diamond$ which does so with at a significantly lower over-all accuracy. Whilst the data for the gender task is still unbalanced, we observe that the na\"{\i}ve classifier can already achieve a better level of fairness leading us to believe this is a fairer task. For this task, BASE is competitive and achieves the second best accuracy and fairness. MFD achieves the greater fairness, however, this comes at the expense of a lower over-all accuracy. In the CelebA scenario BASE achieves the highest performance in all metrics. Again the fairness of the na\"{\i} classifier is low for this scenario, showing that the CelebA task is unfair. These experiments show that BASE works best in an environment that is particularly unfair.

\setlength{\tabcolsep}{4pt}
\begin{table}[!h]
\begin{center}
\caption{
Comparison of methods on UTKFace dataset with age as the target variable. Best results are \textbf{bold} and second best are \underline{underline}. Results marked $\diamond$ are reported directly from \cite{jung_fair_2021}.
}
\label{table:utkface_age}
\begin{tabular}{lllll}
\hline\noalign{\smallskip}
Method $\qquad\qquad$ & Acc. $\uparrow$ & $\sigma_{\mathrm{Acc}}\downarrow$ & $\mathrm{DEO_{max}}\downarrow$ & $\mathrm{DEO_{avg}}\downarrow$ \\
\noalign{\smallskip}
\hline
\noalign{\smallskip}
Na\"{\i}ve Classifier & $82.5\pm1.5$ & $8.8\pm1.2$ & $45.3\pm2.5$ & $25.8\pm1.3$ \\ 
Na\"{\i}ve Classifier Balanced & $83.3\pm1.1$ & $8.7\pm1.3$ & $43.4\pm3.8$ & $21.9\pm2.2$ \\ \hline
MFD$^\diamond$ \cite{jung_fair_2021} & $74.7\pm0.7$ & - & $\mathbf{28.5\pm1.8}$ & $\underline{17.8\pm1.4}$ \\ 
MFD \cite{jung_fair_2021} & $83.4\pm0.5$ & $\underline{6.6\pm1.3}$ & $32.3\pm3.5$ & $18.3\pm2.0$ \\ 
AD \cite{zhang_mitigating_2018} & $\underline{83.6\pm1.4}$ & $7.3\pm1.4$ & $41.0\pm5.6$ & $21.2\pm3.9$ \\ \hline
BASE \textit{Ours} & $\mathbf{83.8\pm0.6}$  & $\mathbf{5.6\pm0.7}$ & $\underline{29.0\pm2.6}$ & $\mathbf{16.0\pm1.3}$ \\
\hline
\end{tabular}
\end{center}
\end{table}
\setlength{\tabcolsep}{1.4pt}

\setlength{\tabcolsep}{4pt}
\begin{table}[!h]
\begin{center}
\caption{
Comparison of methods on UTKFace dataset with gender as the target variable. Best results are \textbf{bold} and second best are \underline{underline}.
}
\label{table:utkface_gender}
\begin{tabular}{lllll}
\hline\noalign{\smallskip}
Method $\qquad\qquad$ & Acc. $\uparrow$ & $\sigma_{\mathrm{Acc.}}\downarrow$ & $\mathrm{DEO_{max}}\downarrow$ & $\mathrm{DEO_{avg}}\downarrow$ \\
\noalign{\smallskip}
\hline
\noalign{\smallskip}
Na\"{\i}ve Classifier & $93.1\pm0.7$ & $2.8\pm0.3$ & $10.3\pm3.2$ & $7.2\pm0.8$ \\
Na\"{\i}ve Classifier Balanced & $\underline{93.5\pm0.9}$ & $2.8\pm0.6$ & $10.0\pm1.5$ & $7.2\pm1.3$ \\\hline
MFD \cite{jung_fair_2021} & $92.4\pm0.4$ & $\mathbf{2.1\pm0.4}$ & $\mathbf{8.0\pm1.0}$ & $\mathbf{5.7\pm0.6}$ \\ 
AD \cite{zhang_mitigating_2018} & $\mathbf{93.9\pm1.1}$ & $2.6\pm0.9$ & $\mathbf{8.0\pm2.6}$ & $\underline{6.2\pm2.0}$ \\ \hline
BASE \textit{Ours} & $93.4\pm0.3$ & $\underline{2.3\pm0.9}$ & $\underline{9.0\pm1.0}$ & $\underline{6.2\pm0.8}$ \\
\hline
\end{tabular}
\end{center}
\end{table}
\setlength{\tabcolsep}{1.4pt}

\setlength{\tabcolsep}{4pt}
\begin{table}[!h]
\begin{center}
\caption{
Comparison of methods on CelebA dataset with attractive as the target variable. Best results are \textbf{bold} and second best are \underline{underline}. Results marked $\diamond$ are reported directly from \cite{park_fair_2022}.
}
\label{table:celeba_attractive}
\begin{tabular}{lllll}
\hline\noalign{\smallskip}
Method $\qquad\qquad$ & Acc. $\uparrow$ & $\sigma_{\mathrm{Acc}}\downarrow$ & $\mathrm{DEO_{max}}\downarrow$ & $\mathrm{DEO_{avg}}\downarrow$ \\
\noalign{\smallskip}
\hline
\noalign{\smallskip}
Na\"{\i}ve Classifier & $79.6\pm0.2$ & $3.0\pm3.5$ & $26.3\pm0.4$ & $25.9\pm0.6$ \\ 
Na\"{\i}ve Classifier Balanced & $\underline{80.1\pm0.2}$ & $\underline{1.1\pm0.3}$ & $\underline{3.5\pm0.4}$ & $\underline{2.1\pm0.4}$ \\ \hline
MFD$^\diamond$ \cite{jung_fair_2021} & $78\pm0.3$ & - & - & $7.4\pm0.3$ \\ 
FSCL$^\diamond$ \cite{park_fair_2022} & $79.1\pm0.4$ & - & - & $11.5\pm0.3$ \\
FSCL+$^\diamond$ \cite{park_fair_2022} & $79.1\pm0.4$ & - & - & $6.5\pm0.4$ \\ \hline
BASE \textit{Ours} & $\mathbf{80.7\pm0.1}$ & $\mathbf{0.8\pm0.2}$ & $\mathbf{3.0\pm0.7}$ & $\mathbf{1.9\pm0.5}$ \\
\hline
\end{tabular}
\end{center}
\end{table}
\setlength{\tabcolsep}{1.4pt}

\subsubsection{Balanced data} \label{sssec:balanced}
Next, we test the scenario in which the training data for the task has been collected with a focus on ensuring that it is balanced with respect to the protected attribute. For this experiment, we use the Fairface dataset and \emph{race} as the protected attribute. For the classification target attribute we test both age and gender. The results are shown in \cref{table:fairface_age,table:fairface_gender}, respectively.

\setlength{\tabcolsep}{4pt}
\begin{table}[!h]
\begin{center}
\caption{
Comparison of methods on Fairface dataset with age as the target variable. Best results are \textbf{bold} and second best are \underline{underline}.
}
\label{table:fairface_age}
\begin{tabular}{lllll}
\hline\noalign{\smallskip}
Method $\qquad\qquad$ & Acc. $\uparrow$ & $\sigma_{\mathrm{Acc.}}\downarrow$ & $\mathrm{DEO_{max}}\downarrow$ & $\mathrm{DEO_{avg}}\downarrow$ \\
\noalign{\smallskip}
\hline
\noalign{\smallskip}
Na\"{\i}ve Classifier & $67.4\pm0.2$ & $2.1\pm0.5$ & $23.6\pm4.3$ & $16.2\pm0.4$ \\
Na\"{\i}ve Classifier Balanced & $66.2\pm0.4$ & $\mathbf{1.9\pm0.4}$ & $\mathbf{12.4\pm0.8}$ & $\mathbf{10.3\pm0.2}$ \\\hline
MFD \cite{jung_fair_2021} & $\underline{68.3\pm0.3}$ & $\mathbf{1.9\pm0.2}$ & $\underline{14.6\pm0.9}$ & $\underline{10.6\pm1.4}$ \\ 
AD \cite{zhang_mitigating_2018} & $\mathbf{68.4\pm0.2}$ & $2.2\pm0.2$ & $21.3\pm0.5$ & $15.7\pm0.9$ \\ \hline
BASE \textit{Ours} & $\mathbf{68.4\pm0.4}$ & $\underline{2.0\pm0.04}$ & $14.8\pm0.5$ & $10.8\pm1.2$ \\
\hline
\end{tabular}
\end{center}
\end{table}
\setlength{\tabcolsep}{1.4pt}

\setlength{\tabcolsep}{4pt}
\begin{table}[!h]
\begin{center}
\caption{
Comparison of methods on Fairface dataset with gender as the target variable. Best results are \textbf{bold} and second best are \underline{underline}.
}
\label{table:fairface_gender}
\begin{tabular}{lllll}
\hline\noalign{\smallskip}
Method $\qquad\qquad$ & Acc. $\uparrow$ & $\sigma_{\mathrm{Acc.}}\downarrow$ & $\mathrm{DEO_{max}}\downarrow$ & $\mathrm{DEO_{avg}}\downarrow$ \\
\noalign{\smallskip}
\hline
\noalign{\smallskip}
Na\"{\i}ve Classifier & $93.4\pm0.05$ & $\underline{2.0\pm1.6}$ & $7.5\pm1.7$ & $\mathbf{6.3\pm0.8}$ \\
Na\"{\i}ve Classifier Balanced & $\mathbf{93.6\pm0.05}$ & $\mathbf{1.9\pm0.2}$ & $\mathbf{6.9\pm0.3}$ & $\underline{6.4\pm0.3}$\\\hline
MFD \cite{jung_fair_2021} & $93.4\pm0.1$ & $2.2\pm0.05$ & $7.7\pm1.1$ & $7.0\pm0.4$ \\ 
AD \cite{zhang_mitigating_2018} & $\mathbf{93.6\pm0.2}$ & $2.1\pm0.1$ & $7.3\pm0.6$ & $6.7\pm0.4$ \\ \hline
BASE \textit{Ours} & $\underline{93.5\pm0.02}$ & $\underline{2.0\pm0.1}$ & $\underline{7.0\pm0.6}$ & $\underline{6.4\pm0.4}$ \\
\hline
\end{tabular}
\end{center}
\end{table}
\setlength{\tabcolsep}{1.4pt}

In these two scenarios, we observe that the fairness of the na\"{\i}ve classifier is already high due to the balanced nature of the data. For both target attributes, the na\"{\i}ve classifier with balanced sampling achieves the best fairness for two of the three metrics. However, this comes at the cost of accuracy for the age task. For both tasks BASE achieves the second best results for $\sigma_{\mathrm{Acc.}}$, with equal highest overall accuracy in the age task and the second best overall accuracy for the gender task.

\subsubsection{Biased data}\label{sssec:biased_data}

\begin{figure}[!h]
\centering
\includegraphics[width=0.8\textwidth]{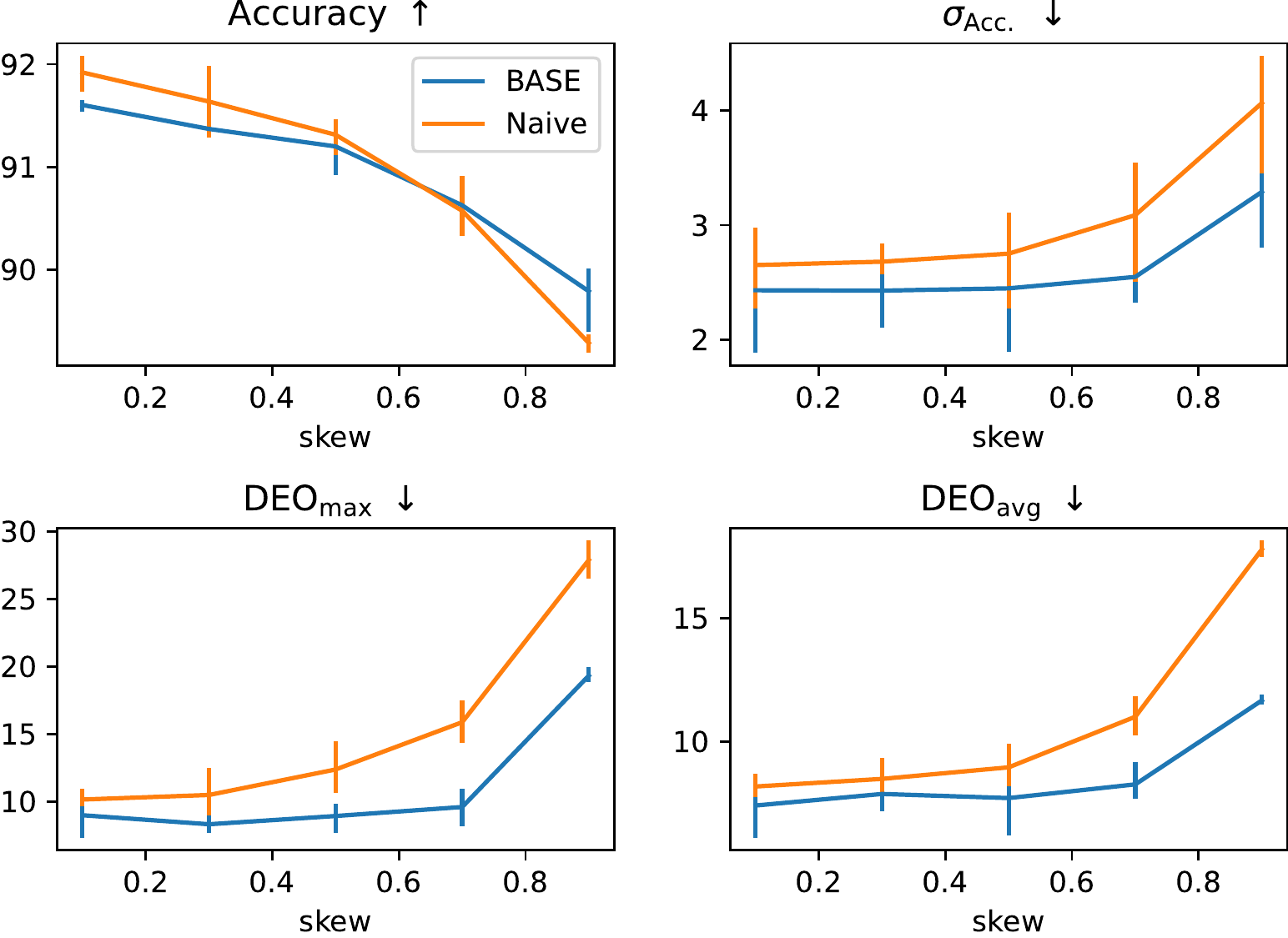}
\caption{
The accuracy and fairness of a Na\"{\i}ve classifier and BASE over different skew parameters of the FairfaceS dataset. Error bars are the 95\% confidence interval over 3-fold cross-validation.
}
\label{fig:skew_results}
\end{figure}

Finally, we investigate how out method performs with an increasingly biased dataset. For this experiment we use the FairfaceS dataset (\cref{fairfaces}) with gender as the target variable. We evaluate a na\"{\i}ve classifier and BASE over a range of different skew parameters and observe the effect on accuracy and fairness. The results are shown in \cref{fig:skew_results}.

We observe that, as one would expect, as the skew increases and consequentially the bias in the dataset increases both accuracy and fairness decay for both methods. Additionally, at low levels of bias, whilst BASE is able to increase the fairness of the classifier in all metrics, this comes at the cost of overall accuracy compared to the na\"{\i}ve classifier. However, as the skew increases the accuracy of the na\"{\i}ve classifier decays at a greater rate than BASE. At extreme skew levels, BASE is even able to achieve a higher degree of overall accuracy. The same results can be seen with the fairness metrics. With the performance of the na\"{\i}ve classifier decaying at a higher rate than BASE. Even though BASE produces a more fair predictor at low skew levels, the performance gap only increases as the skew increases.
\section{Conclusion}
In this work, we introduce a new fairness objective based upon optimising the standard deviation of soft accuracy across demographics of a protected attribute. Experimental results on UTKFace, CelebA, Fairface and FairfaceS show that our system is able to produce fairer AI models for computer vision tasks under widely varying conditions whilst being particular effective for more unfair scenarios and can even improve the overall accuracy compared to a naive model in heavily biased data-sets. 

\bibliographystyle{splncs04}
\bibliography{references}

\appendix
\section{Training Procedure}\label{train_proc}
All models were trained with the optimization hyper-parameters described in \cref{opt_param}. As a default we used a batchsize of 512, a learning rate of 0.0001 and a cosine learning rate schedule \cite{loshchilov_sgdr_2017}. However, in some cases we found extra performance could be found by varying these hyper-parameters. The exact values that were used for each experiment is shown in \cref{opt_param_all}. Model weights were initialised using models from \cite{wightman_resnet_2021}, which are pretrained upon the ImageNet data-set \cite{deng_imagenet_2009}.

\begin{table}[!ht]
\centering
\caption{{\bf Optimization hyper-parameters used during training.}}
\begin{tabular}{|l l|}
\hline
Parameter & Value \\ \hline \hline
Optimizer & AdamW \cite{loshchilov_decoupled_2019} \\ 
Weight Decay & 0.02 \\
Epochs & 50 \\

\hline
\end{tabular}
\label{opt_param}
\end{table}

\begin{table}[!ht]
\centering
\caption{{\bf Learning rate and batch size used during training.}}
\begin{tabular}{|l l l l l l l|}
\hline
Dataset & Task & Method & Batchsize & Learning Rate & Schedule & Image Size \\ \hline \hline
UTKFace & Age & Naive & 512 & 0.0001 & Cosine & 176 \\
UTKFace & Age & Naive Balanced & 512 & 0.0001 & Cosine & 176 \\
UTKFace & Age & MFD & 128 & 0.001 & None & 176 \\
UTKFace & Age & AD & 128 & 0.0001 & Cosine & 176 \\
UTKFace & Age & BASE & 512 & 0.0001 & Cosine & 176 \\ \hline
UTKFace & Gender & Naive & 512 & 0.0001 & Cosine & 176 \\
UTKFace & Gender & Naive Balanced & 512 & 0.0001 & Cosine & 176 \\
UTKFace & Gender & MFD & 512 & 0.0001 & Cosine & 176 \\
UTKFace & Gender & AD & 128 & 0.0001 & Cosine & 176 \\
UTKFace & Gender & BASE & 512 & 0.0001 & Cosine & 176 \\ \hline
CelebA & Gender & Naive & 512 & 0.0001 & Cosine & 128 \\
CelebA & Gender & Naive Balanced & 512 & 0.0001 & Cosine & 128 \\
CelebA & Gender & BASE & 512 & 0.0001 & Cosine & 128 \\ \hline
Fairface & Age & Naive & 512 & 0.0001 & Cosine & 176 \\
Fairface & Age & Naive Balanced & 512 & 0.0001 & Cosine & 176 \\
Fairface & Age & MFD & 512 & 0.0001 & Cosine & 176 \\
Fairface & Age & AD & 128 & 0.0001 & Cosine & 176 \\
Fairface & Age & BASE & 512 & 0.0001 & Cosine & 176 \\ \hline
Fairface & Gender & Naive & 512 & 0.0001 & Cosine & 176 \\
Fairface & Gender & Naive Balanced & 512 & 0.0001 & Cosine & 176 \\
Fairface & Gender & MFD & 512 & 0.0001 & Cosine & 176 \\
Fairface & Gender & AD & 128 & 0.0001 & Cosine & 176 \\
Fairface & Gender & BASE & 512 & 0.0001 & Cosine & 176 \\ \hline
FairfaceS & Gender & Naive & 512 & 0.0001 & Cosine & 176 \\
FairfaceS & Gender & BASE & 512 & 0.0001 & Cosine & 176 \\ \hline

\hline
\end{tabular}
\label{opt_param_all}
\end{table}

Data augmentation was also used to assist in regularization of the models. The data augmentation scheme, shown in \cref{data_aug}, was used for all methods and tasks. At inference time the images were resized, and then centre cropped. 

\begin{table}[!ht]
\centering
\caption{{\bf Data augmentation transformations used during training.}}
\begin{tabular}{|l|}
\hline
Transformation \\ \hline
Horizontal Flip ($p=0.5$) \\ 
Random Resized Crop (Scale $[0.08,1]$, Ratio $[0.75,1.33]$) \\
Color Jitter (factor $0.4$, $p=0.8$) \\
\hline
\end{tabular}
\label{data_aug}
\end{table}
\end{document}